\documentclass[letterpaper]{article} 
\usepackage{aaai23}  
\usepackage{times}  
\usepackage{helvet}  
\usepackage{courier}  
\usepackage[hyphens]{url}  
\usepackage{graphicx} 
\usepackage{natbib}  
\usepackage{caption} 
\usepackage{algorithm}
\usepackage{algorithmic}

\usepackage{booktabs}
\usepackage{amsfonts}       
\usepackage{amsmath}
\usepackage{diagbox}

\usepackage{amsthm}
\usepackage{multirow}

\usepackage{newfloat}
\usepackage{listings}
\DeclareCaptionStyle{ruled}{labelfont=normalfont,labelsep=colon,strut=off} 
\floatstyle{ruled}
\newfloat{listing}{tb}{lst}{}
\floatname{listing}{Listing}
\title{A Scope Sensitive and Result Attentive Model for\\ Multi-Intent Spoken Language Understanding}
\author{
    Lizhi Cheng\textsuperscript{\rm 1},
    Wenmian Yang\textsuperscript{\rm 2}{$^*$},
    Weijia Jia\textsuperscript{\rm 3}\thanks{Corresponding authors.}\\
}
\affiliations{
    \textsuperscript{\rm 1}Shanghai Jiao Tong University, Shanghai, PR China\\
    \textsuperscript{\rm 2}Nanyang Technological University, Singapore, Singapore\\
    \textsuperscript{\rm 3}BNU-UIC Institute of Artificial Intelligence and Future Networks, Beijing Normal University (Zhuhai),
    Guangdong Key Lab of AI and Multi-Modal Data Processing,
    BNU-HKBU United International College, Zhuhai, Guang Dong, PR China\\

    clz19960630@sjtu.edu.cn, 
    wenmian.yang@ntu.edu.sg,
    jiawj@bnu.edu.cn
}

\usepackage{bibentry}

\begin{document}
\maketitle

\begin{abstract}
Multi-Intent Spoken Language Understanding (SLU), a novel and more complex scenario of SLU, is attracting increasing attention. 
Unlike traditional SLU, each intent in this scenario has its specific scope. Semantic information outside the scope even hinders the prediction, which tremendously increases the difficulty of intent detection. More seriously, guiding slot filling with these inaccurate intent labels suffers error propagation problems, resulting in unsatisfied overall performance. 
To solve these challenges, in this paper, we propose a novel Scope-Sensitive Result Attention Network (SSRAN) based on Transformer, which contains a Scope Recognizer (SR) and a Result Attention Network (RAN). 
Scope Recognizer assignments scope information to each token, reducing the distraction of out-of-scope tokens. Result Attention Network effectively utilizes the bidirectional interaction between results of slot filling and intent detection, mitigating the error propagation problem. 
Experiments on two public datasets indicate that our model significantly improves SLU performance (5.4\% and 2.1\% on Overall accuracy) over the state-of-the-art baseline.
\end{abstract}

\section{Introduction}
\label{sec:introduction}
\begin{figure}[t]
\centering
\includegraphics[width=0.48\textwidth]{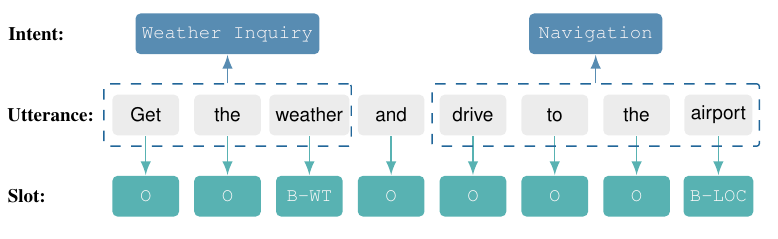}
\caption{An example of Multi-Intent SLU, where \texttt{B-WT} denotes \texttt{B-Weather}, and \texttt{B-LOC} denotes \texttt{B-Location}.}
\label{fig:example}
\end{figure}

Spoken Language Understanding (SLU) is a core component of task-oriented dialogue systems, generally containing two subtasks, namely Slot Filling (SF) and Intent Detection (ID) \cite{SLU2011}. In traditional SLU tasks, SF is a sequence labeling task aiming to fill in the slot for each token; ID is a sentence-level semantic classification task that identifies the intent label for the entire utterance. Recent studies \cite{gangadharaiah2019joint,qin2020agif} find that users also express more than one intent in an utterance in many scenarios. Thus, a new SLU task, i.e., Multi-Intent SLU, is derived, attracting increasing attention. A simple example of Multi-Intent SLU is shown in Figure \ref{fig:example}.

In Multi-Intent SLU, the input utterance is usually composed of multiple sub-utterances, and each sub-utterance has its own intent. The final ID results are the union of the intent of all sub-utterances. Moreover, the intent is usually closely related to the tokens within its sub-utterance and has a weak relationship with the others.
For example, as shown in Figure \ref{fig:example}, the intent \texttt{Weather Inquiry} of the previous sub-utterance is related to the token \texttt{Get} and \texttt{weather}. And the intent of the second sub-utterance is mainly originated from token \texttt{drive} and \texttt{airport}, but not \texttt{weather}. Thus, in the Multi-Intent SLU task, each intent has its specific scope, and the semantic information outside the scope even negatively affects the ID prediction. Due to the above, the difficulty of ID greatly increased compared with traditional SLU.
Therefore, how to divide the scope for each intent effectively to enhance the performance of ID is a major challenge.

Moreover, recent studies \cite{qin2019stack,clzICME2021,cheng2021effective} find that ID and SF are closely related, and ID results positively promote the predicting performance of SF.
\citet{qin2021glgin,qin2020agif} extend this idea to Multi-Intent SLU and try to utilize the ID results to improve SF via the Graph Attention mechanism. However, with the increase in difficulty, the accuracy of ID in Multi-Intent SLU is not as satisfactory as in traditional SLU.
As a result, employing ID results to guide SF will inevitably cause the error propagation problem, affecting SF performance and reducing the overall accuracy of the model. 
Therefore, how to mitigate the error propagation problem and effectively utilize the correlation between ID and SF to improve the final performance is also a significant challenge.

To solve the above challenges, in this paper, we propose a Scope Sensitive Result Attention Network (SSRAN). Our SSRAN contains a Scope Recognizer (SR) and a Result Attention bi-feedback Network (RAN) based on Transformer. The SR is proposed to capture the scope information for the first challenge. In practice, we first predict preliminary results of ID and SF by the encoder output and obtain their result-embedding vectors.
Then we calculate the scope-weighted matrix according to the impact of these results on each token and obtain the scope-sensitive hidden states and result-embedding vectors with this matrix. The SR incorporates scope information into the model and reduces the distraction of out-of-scope tokens.

To further solve the error propagation problem for the second challenge, we design RAN based on the self-attention mechanism. 
RAN fuses the semantics information of ID and SF results based on their interdependencies (i.e., the semantic similarities of each other) and returns a comprehensive result-semantic vector. We merge the result-semantic vector with the scope-sensitive hidden states to help the following prediction. By RAN, we refine ID and SF bidirectionally, which mitigates the error propagation and overcomes the weakness of one-way refinement (i.e., only ID to SF).
Moreover, we design two auxiliary tasks called Intent Number Prediction (INP) and Slot Chunking Task (SCT) for multi-task learning, aiming to improve the performance of ID and SF, respectively.

The main contributions of this paper are presented as follows:
\begin{enumerate}
\item We propose SR to enhance the ID accuracy of Multi-Intent SLU, which focuses on the scope of intent and reduces the distraction of out-of-scope tokens.
\item We mitigate the error propagation problem caused by mistaken ID labels to improve overall performance by our RAN, which conducts bidirectional interaction between ID and SF to improve both subtasks.
\item Experimental results on two public datasets show our model is superior to existing SOTA models in terms of all evaluation metrics.
\end{enumerate}

\section{Related Work}
\label{sec: Related Work}
In this section, we introduce the related work from two aspects.

\subsection{Slot Filling and Intent Detection}  
Early studies of SLU \cite{yao2014lstm,mesnil2014rnn,peng2015rnn,kurata2016lstm} prove that predicting SF and ID separately with independent models is less effective than the joint models. Motivated by the above, \citet{Slot-gated2018} firstly propose a gate mechanism to learn the relationship between slot and intent. Inspired by \citet{Slot-gated2018}, \citet{bi-dictional2019,liu2019cm,JointCapsule2019} dive deeper into the relationship between ID and SF and propose some bi-directional networks. The above works mainly focus on the hidden states of the two tasks and lack the result information. Then, \citet{qin2019stack} propose a Stack-Propagation framework to utilize the result information, which refines SF with the ID label. Considering the impact of SF results on the ID task, inspired by \citet{yang2019}, \citet{clzICME2021} propose a bi-feedback network RPFSLU, which guides the second round prediction by the first round results via representing learning. Although RPFSLU performs satisfied, it surfers an extended inference latency caused by predicting SLU results in multiple rounds. To accelerate the inferring process, \citet{wu2020slotrefine} propose a non-autoregressive SlotRefine based on Transformer, which successfully speeds up but encounters the uncoordinated-slot problem. To handle this problem, \citet{cheng2021effective} propose LR-Transformer with a Layer Refined Mechanism and a specially designed auxiliary task. \citet{clzTOIS} then expend this work to multi-turn SLU tasks with a Salient History Attention module.

\subsection{Multi-Intent SLU}
The models above are designed based on the assumption that each utterance only has one single intent. Focusing on the multiple intents scenario, \citet{xu2013convolutional}
and \citet{kim2017two} begin to study the Multi-Intent SLU. \citet{gangadharaiah2019joint} jointly learning SF and multiple ID via a multi-task framework. \citet{qin2020agif} extend their idea in traditional SLU and utilize ID labels in SF with an adaptive interaction network. They upgrade their framework to a non-autoregressive model in \cite{qin2021glgin} and achieve the SOTA performance.
However, their models are at risk of error propagation due to the increasing difficulty of multiple ID prediction.
Compared with their models, we focus on the scope of the intent and reduce the distraction of out-of-scope tokens. Moreover, we also incorporate the impact of SF results on the ID task into the model. These modifications effectively enhance ID accuracy, mitigate error propagation, and improve overall performance.

\section{Method}

\begin{figure*}[t]
\centering
\includegraphics[width=0.95\textwidth]{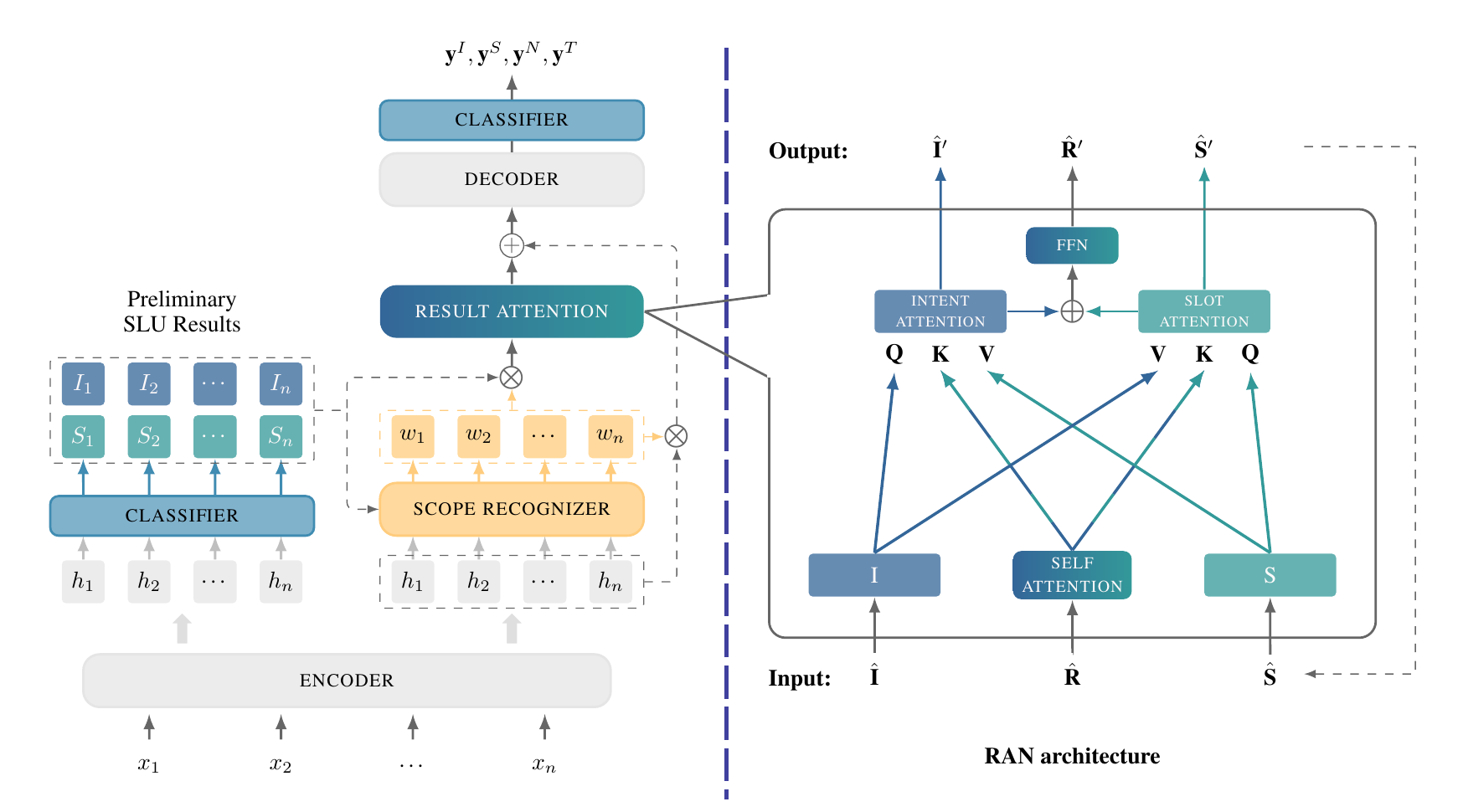}
\caption{General framework.}
\label{fig:model}
\end{figure*}

\subsection{Problem Formulation}
In this section, we introduce the problem formulation for the Multi-Intent SLU task.

The input of SLU tasks is an utterance composed by a token sequence $\textbf{X}=\{x_1,...,x_n\}$, where $n$ denotes the sequence length.
Given $\textbf{X}$ as input, our tasks are composed of Slot Filling (SF) and Intent Detection (ID).
Specifically, SF is a sequence labeling task to predict a slot label sequence $\textbf{y}^S=\{\textbf{y}^S_1,...,\textbf{y}^S_n\}$,
while ID is a multi-label semantic classification task to predict the intent labels $\textbf{y}^I=\{\textbf{y}^I_1,...,\textbf{y}^I_m\}$ for the whole utterance, where $m$ denotes the number of intents in given utterance. 

\subsection{Overview}
In this section, we describe the overview of SSRAN and introduce the relationship between each module.

The general framework of our SSRAN is shown in Figure \ref{fig:model}, which consists of four parts: an encoder, a Span Recognizer (SR), a Result Attention Network (RAN), and a decoder.
The encoder and decoder are based on the Transformer Encoder Layer \cite{Attention_is_all_you_need} with relative position representations \cite{shaw2018position_relative_attention}; SR and RAN work between the encoder and the decoder, which are the core components of SSRAN.

In practice, we first model the input utterance into a hidden states sequence by our encoder. Then, we get the preliminary SF and ID results according to the hidden states and embed these results. 
Next, we employ our SR and RAN to incorporate scope and result information into the model and guide the final prediction via the decoder.

More specifically, in SR, we calculate a scope-weighted matrix, with which we obtain the scope-sensitive hidden states and result embedding vectors.
Subsequently, in RAN, we fuse the semantics information of ID and SF results based on their interdependencies via the attention mechanism and return a result-semantic vector. 
Finally, we merge the result-semantic vector and the scope-sensitive hidden states and obtain the SLU output through our decoder.

In addition, we design two auxiliary tasks,i.e., Intent Number Prediction (INP) and Slot Chunking Task (SCT), for multi-task learning. For INP, we predict the number of intents of each utterance to enhance the ID accuracy. For SCT, we predict the tag of BIO (short for beginning, inside, and outside) of each slot to address sequential dependency.

\subsection{Encoder}
Recent studies prove SLU results contain specific semantic information, which needs to be considered in prediction. Thus, we employ an encoder to predict preliminary results and incorporate the result semantic information into the model.

Specifically, with the input of $\textbf{X}=\{x_1,...,x_n\}$, we first obtain a hidden states sequence $\textbf{H}=\{\textbf{h}_1,...,\textbf{h}_n\ \in \mathbb{R}^{d_{model}} \}$, where $d_{model}$ is Transformer input and output size.
Our encoder is composed of two Transformer Encoder layers. Since the Transformer framework is well known, we do not need to introduce it in every detail. 

Subsequently, we predict the preliminary SLU results $\tilde{\textbf{y}}^S = \{ \tilde{\textbf{y}}^S_1,...,\tilde{\textbf{y}}^S_n \in \mathbb{R}^{d_s} \}$, $\tilde{\textbf{y}}^I = \{ \tilde{\textbf{y}}^I_1,...,\tilde{\textbf{y}}^I_n \in \mathbb{R}^{d_i} \}$ by 
\begin{align}
\begin{aligned}
    \tilde{\textbf{y}}^S_j &=\textbf{W}^S  \left(\textbf{h}_j \oplus \mathrm{Pool} \left(\textbf{H}\right)\right) + \textbf{b}^S\\
    \tilde{\textbf{y}}^I_j &= \textbf{W}^I  \left(\textbf{h}_j \oplus \mathrm{Pool} \left(\textbf{H}\right)\right) + \textbf{b}^I\\
\end{aligned}
\label{eq:getres0}
\end{align}
where 
$\textbf{W}^S \in \mathbb{R}^{ d_{s} \times 2d_{model}}$
and $\textbf{W}^I \in \mathbb{R}^{ d_{i} \times 2d_{model} }$ are fully connected matrices, 
$\textbf{b}^S \in \mathbb{R}^{d_{s}}$  and $\textbf{b}^I \in \mathbb{R}^{d_{i}}$ are bias vectors, 
$d_{s}$ and $d_i$ are the categories of the slot labels and intent labels,
$\oplus$ denotes the concatenation operation, and $\mathrm{Pool}$ denotes average polling operation.

We then embed these results for the bi-feedback process.
Specifically, we obtain SF result embedding vectors $\textbf{S}=\{\textbf{S}_1,...,\textbf{S}_n\ \in \mathbb{R}^{d_{e}} \}$ and ID result embedding vectors $\textbf{I}=\{\textbf{I}_1,...,\textbf{I}_n\ \in \mathbb{R}^{d_{e}} \}$ by
\begin{align}
\begin{aligned}
    \textbf{S}_j &=  \textbf{E}^S \cdot \mathrm{softmax} \left(  \tilde{\textbf{y}}^S_j\right)\\
    \textbf{I}_j &=  \textbf{E}^I \cdot \mathrm{softmax} \left( \tilde{\textbf{y}}^I_j \right)\\
\end{aligned}
\label{eq:getres1}
\end{align}
where 
$\textbf{E}^S \in \mathbb{R}^{ d_{e} \times d_{s} }$ and
$\textbf{E}^I \in \mathbb{R}^{ d_{e} \times d_{i} }$ are fully connected matrices, and $d_e$ is the embedding size.

\subsection{Scope Recognizer}
Each intent in Multi-Intent SLU has its specific scope. In this section, we propose the Scope Recognizer (SR) to effectively divide the scope.

Specifically, we first calculate a scope weight matrix $\textbf{W}=\{ \textbf{w}_1,...,\textbf{w}_n  \}$.
The scope weight of each token $\textbf{w}_j = \{ w_{j,1},...,w_{j,n} \} $ is calculated by
\begin{align}
    &\ \ \ \ \ \ \ \ \ \ \ \ \ \ \ \ w_{j,k}=\frac{\exp \alpha_{j,k}}{\sum_{l=1}^{n} \exp \alpha_{j,l}}\\
    &\alpha_{j,k}=\frac{\left(\textbf{h}_{j} W^{1}\right) \left( ( \textbf{h}_{k}+\textbf{I}_{k}+\textbf{S}_{k}) W^{2}\right)^{T}}{\sqrt{d_{model}}}
\end{align}
where $\textbf{W}^1,\textbf{W}^2 \in \mathbb{R}^{ d_{model} \times d_{model} }$ are fully connected matrices.

We then obtain the scope-sensitive hidden states $\hat{\textbf{H}}=\{\hat{\textbf{h}}_1,...,\hat{\textbf{h}}_n\ \in \mathbb{R}^{d_{model}} \}$ by
\begin{align}
    \hat{\textbf{h}}_j = \textbf{h}_j + \sum_{k=1}^{n}  w_k \textbf{h}_k
\label{eq: ss-hidden}
\end{align}
Similarly, we obtain the scope-sensitive result-embedding vector $\hat{\textbf{S}}$ and $\hat{\textbf{I}}$ by
\begin{align}
    \hat{\textbf{S}}_j &= \textbf{S}_j + \sum_{k=1}^{n}  w_k \textbf{S}_k\\
    \hat{\textbf{I}}_j &= \textbf{I}_j + \sum_{k=1}^{n}  w_k \textbf{I}_k
\label{eq: ss-re}
\end{align}

By our SR, we address the scope information for hidden states and result-embedding vectors, which will be further utilized to enhance the final prediction. 

\subsection{Result Attention Network}
As we mentioned in section \textbf{Introduction}, 
considering ID results only in SF cause the error propagation problem and affects the final prediction. To mitigate this problem, in this section, we propose a Result Attention Network(RAN) to effectively utilize the result information of one task in the other.


As shown in Figure \ref{fig:model}, RAN is a multi-layer architecture (3 layers in practice) and each layer operates in the same way. The input of the first layer is $\hat{\textbf{S}}$, $\hat{\textbf{I}}$, and a initialed result-semantic vector $\hat{\textbf{R}}=\hat{\textbf{I}}+\hat{\textbf{S}}$. And each RAN layer returns processed vectors, i.e., $\hat{\textbf{S}}{'}$, $\hat{\textbf{I}}{'}$, and $\hat{\textbf{R}}{'}$, which are used as the input of the next RAN layer. 

Specifically,
we first address $\hat{\textbf{R}}$ by
\begin{align}
    &\tilde{\textbf{R}} = \operatorname{Attention}(\hat{\textbf{R}},\hat{\textbf{R}},\hat{\textbf{R}})\\
    &\textbf{R}^{att} = \operatorname{Norm}(\hat{\textbf{R}}+\tilde{\textbf{R}})
\label{eq: self-attention2}
\end{align}
where $\operatorname{Attention}(q,k,v)$ is the self-attention function and $\mathrm{Norm}(x)$ is the layer-normalization function used in Transformer \cite{Attention_is_all_you_need}. 

To mitigate the error propagation problem, we conduct two cross-attention operations between SF and ID to incorporate salient semantic information from both two tasks by
\begin{align}
    &\tilde{\textbf{S}}\ = \operatorname{Attention}(\hat{\textbf{S}} ,\textbf{R}^{att},\hat{\textbf{I}})\\
    &\hat{\textbf{S}}{'} = \operatorname{Norm}(\hat{\textbf{S}}+\tilde{\textbf{S}})\\
    &\tilde{\textbf{I}}\ = \operatorname{Attention}(\hat{\textbf{I}} ,\textbf{R}^{att},\hat{\textbf{S}})\\
    &\hat{\textbf{I}}{'} = \operatorname{Norm}(\hat{\textbf{I}}+\tilde{\textbf{I}})
\label{eq: self-attention3}
\end{align}
Through this operation, we update the SF and ID result-embedding vectors according to the result semantic information from the opposite task.

Finally, we further utilize these two vectors to update the result-semantic vector $\tilde{\textbf{R}}{'} =\hat{\textbf{S}}{'} + \hat{\textbf{I}}{'}$ 
and prevent the vanishing or exploding of gradients by
\begin{align}
    \hat{\textbf{R}}{'} &= \operatorname{Norm}(\tilde{\textbf{R}}{'}+\operatorname{FFN}(\tilde{\textbf{R}}{'}))
\end{align}
where FFN is a feed-forward network consisting of two linear transformations with a ReLU activation function in between.

RAN finally returns a result-semantic vector $\textbf{R}=\{\textbf{R}_1,...,\textbf{R}_n \in \mathbb{R}^{d_{model}}\}$ from the output of the last RAN layer, which contains both semantic information in SF results and ID results.

\subsection{Decoder}

With $\textbf{R}$ and scope sensitive hidden states $\hat{\textbf{H}}$, we obtain a comprehensive hidden states sequence $\textbf{H}^{rs} = \{\textbf{h}^{rs}_1,...,\textbf{h}^{rs}_n \in \mathbb{R}^{d_{model}}\} $ by
\begin{align}
    \textbf{H}^{rs} &= \mathrm{Norm} ( \hat{\textbf{H}} + \textbf{R} + \mathrm{FFN} (\mathrm{Pool}(\hat{\textbf{H}})) )
\label{Eq: merge}
\end{align}
By this operation, $\textbf{H}^{rs}$ is both \textbf{r}esult-semantic-attentive and \textbf{s}cope-sensitive.

Then we obtain a hidden states sequence $\textbf{H}^d=\{\textbf{h}^d_1,...,\textbf{h}^d_n\ \in \mathbb{R}^{d_{model}} \}$ through our decoder, which is composed of four Transformer Encoder layers. 

Finally we predict the SF result and ID result of each token by
\begin{align}
\begin{aligned}
    \hat{\textbf{y}}^S_j &=  \textbf{W}^S  \left(\textbf{h}^d_j \oplus \mathrm{Pool} (\textbf{H}^d)\right) + \textbf{b}^S \\
    \hat{\textbf{y}}^I_j &=  \textbf{W}^I  \left(\textbf{h}^d_j \oplus \mathrm{Pool} (\textbf{H}^d)\right) + \textbf{b}^I \\
\end{aligned}
\label{eq:getres2}
\end{align}

Notably, we utilize the same classifier to predict the preliminary result and the final result.

The joint loss function of SLU is defined as:
\begin{align}
    &\mathcal{L}_{SF} = -\sum_{j=1}^{n} 
        \textbf{y}^S_j\log \left( \mathrm{softmax}(\hat{\textbf{y}}^S_j) \right)\\
    &\mathcal{L}_{ID} = -\sum_{j=1}^{n} 
        \mathrm{BCE} \left(\textbf{y}^I_j,\sigma(\hat{\textbf{y}}^I_j) \right)\\
    &\mathcal{L}_{SLU} = \alpha \mathcal{L}_{SF} + (1-\alpha) \mathcal{L}_{ID}
\label{eq: sluloss}
\end{align}
where ${\textbf{y}}^{S}_j$ and ${\textbf{y}}^{I}_j$ are the one-hot vector of ground truth labels, $\alpha$ is a hyper-parameter,
$\sigma$ is the $\mathrm{sigmoid}$ function, 
and $\mathrm{BCE}$ is defined as:
\begin{equation}
    \mathrm{BCE}(\textbf{y},\hat{\textbf{y}})=\textbf{y} \log (\hat{\textbf{y}})+(1-\textbf{y}) \log (1-\hat{\textbf{y}})
\end{equation}

During inference, we obtain the label of SF by
\begin{equation}
\begin{aligned}
    o^S_j &= \mathrm{argmax} ( \hat{\textbf{y}}^S_j)
\end{aligned}
\label{eq: getS}
\end{equation}
and the label of ID by
\begin{equation}
    o^I = \mathrm{Topk} \left( \sum_{j=1}^{N} \mathrm{softmax} ( \hat{\textbf{y}}^I_j)\right)
\label{eq: getI}
\end{equation}
where $\mathrm{Topk}$ indicates to select $k$ labels with the top probabilities.
We determine $k$ by our INP task that will be introduced in the next section.

\subsection{Auxiliary Task}
In this section, we design Intent Number Prediction (INP) and Slot Chunking
Task (SCT) to further enhance the SLU performance.


Specifically, we predict the INP result $\hat{\textbf{y}}^N \in \mathbb{R}^{d_N}$ with our decoder output $\textbf{H}^d$ by
\begin{align}
    \hat{\textbf{y}}^N &=   \textbf{W}^{N} \cdot \mathrm{Pool} (\textbf{H}^d) + \textbf{b}^{N}
\label{eq: getN}
\end{align}
where $\textbf{W}^N \in \mathbb{R}^{ d_{N} \times d_{model}}$
is a fully connected matrix, 
$\textbf{b}^N \in \mathbb{R}^{d_{N}}$ is a bias vector, 
and $d_{N}$ is equal to $d_{i}$.

And the loss function of INP is defined as:
\begin{equation}
    \mathcal{L}_{INP} =-\textbf{y}^{N} \log \left( \mathrm{softmax}(\hat{\textbf{y}}^N) \right)
\end{equation}
where $\textbf{y}^{N}$ is the one-hot vector of the ground truth label of INP.

During inference, we determine the parameter $k$ in Eq.\ref{eq: getI} by
\begin{equation}
    k = \arg\max(\hat{\textbf{y}}^N)
\end{equation}




Besides, previous studies \cite{wu2020slotrefine,cheng2021effective} find that SF suffers the uncoordinated slot problem when using Transformer based models. 
Since SF utilizes the ``Beginning-Inside–Outside (BIO)" tagging format, which clearly divides the slot chunk, we design SCT to attain sequential dependency and enhance SF performance according to the BIO tagging. 

In practice, we predict the SCT result $\hat{\textbf{y}}^T = \{ \hat{\textbf{y}}^T_1,...,\hat{\textbf{y}}^T_n \in \mathbb{R}^{d_N}\} $ with our decoder output $\textbf{H}^d$ by
\begin{align}
    \hat{\textbf{y}}^T_j &=   \mathrm{softmax}(\textbf{W}^{T} \textbf{H}^d_j + \textbf{b}^{T})
\label{eq: getT}
\end{align}
And the loss function of SCT is defined as:
\begin{equation}
    \mathcal{L}_{SCT} =- \sum_{j=1}^{n} \textbf{y}^{T}_j \log \left(\mathrm{softmax}( \hat{\textbf{y}}^T_j) \right)
\end{equation}
where $\textbf{y}^{T}$ is is the one-hot vector of the ground truth label, i.e., one of \texttt{O},\texttt{B-Tag} and \texttt{I-tag}.

We utilize INP and SCT for multi-task learning and the joint loss function of our whole task is
\begin{equation}
    \mathcal{L} = \mathcal{L}_{SLU} + \lambda (\mathcal{L}_{INP} + \mathcal{L}_{SCT})
\label{eq: totalloss}
\end{equation}
where $\lambda$ is a hyper-parameter.

\section{Experiment}
In this section, we first introduce the experiment setup. 
Then, we show the experiment results and conduct ablation studies. 
Next, we provide a case study and visualization.
Finally, we analyze the effect of pre-trained models on SSRAN.

\subsection{Experimental Settings and Baselines}
\begin{table}[h]
\centering
\begin{tabular}{lrrr}
\toprule
Dataset                 & MixATIS      & MixSNIPS     \\
\midrule
Vocabulary Size         & 722       & 11241     \\
Intent categories             & 17        & 6         \\
Slot categories              & 116       & 71        \\
Training set size            & 13162      & 39776     \\
Validation set size          & 756       & 2198       \\
Test set size                 & 828       & 2199       \\
\bottomrule
\end{tabular}
\caption{Dataset statistics.}
\label{dataset}
\end{table}

\paragraph{Dataset:}
To evaluate the efficiency of our proposed model, we conduct experiments on two public datasets, i.e., MixATIS and MixSNIPS \cite{atis,snips,qin2020agif}.
The dataset statistics are shown in Table \ref{dataset}.

\paragraph{Evaluation Metrics:}
Following previous work, we evaluate the SLU performance of ID by accuracy and the performance of SF by the F1 score. Besides, we utilize overall accuracy to indicate the proportion of utterances whose slots and intents are both correctly predicted. 

\paragraph{Set up:}
Following previous work, we use Adam \cite{adam} to optimize the parameters in our model and adopt the suggested learning rate of 0.001. The batch size is set to 32 according to the training data size. We choose Transformer input and output size $d_{model}$ as 128, the size of the inner-layer in the feed-forward network $d_{ff}$ as 512, the number of attention heads as 8, and the dropout ratio as 0.1.
The hyper-parameter $\alpha$ used in Eq.\ref{eq: sluloss} is set as 0.65, and $\lambda$ in Eq.\ref{eq: totalloss} is set as 0.3, respectively. When tuning hyper-parameters, we repeat the model 5 times and select the parameters with the best average performance on the validation set as the optimal.

\begin{table*}[t]
\centering
\begin{tabular}{l|ccc|ccc}
\hline
\multirow{2}{*}{\textbf{Model}}  & \multicolumn{3}{c|}{\textbf{MixATIS}}    & \multicolumn{3}{c}{\textbf{MixSNIPS}} \\ 
\cline{2-7} & \multicolumn{1}{c}{Intent} & \multicolumn{1}{c}{Slot} & Overall & \multicolumn{1}{c}{Intent} & \multicolumn{1}{c}{Slot} & Overall \\ 
\hline
Stack-Propagation \cite{qin2019stack} & 72.1  & 87.8  & 40.1    & 96.0  & 94.2  & 72.9     \\
Joint Multiple ID-SF \cite{gangadharaiah2019joint}   & 73.4       & 84.6          & 36.1       & 95.1       & 90.6         & 62.9   \\
AG-IF \cite{qin2020agif}    & 74.4       & 86.7          & 40.8        & 95.1       & 94.2          & 74.2   \\
GL-GIN \cite{qin2021glgin}      & 76.3       & 88.3          & 43.5        & 95.6       & 94.9          & 75.4   \\
LR-Transformer \cite{cheng2021effective}           & 76.1      & 88.0          & 43.3        & 95.6       & 94.4          & 74.9   \\
\hline
Basic model  & 75.1      & 85.2      & 41.2      & 95.2      & 94.2      & 73.2 \\
SSRAN w/o SR & 76.5      & 88.7      & 46.3      & 96.9      & 95.4      & 76.7 \\
SSRAN w/o RAN & 76.9     & 87.6      & 43.0      & 96.9      & 93.9     & 75.1 \\
SSRAN w/o Auxiliary Tasks & 77.1      & 89.1     & 47.2      & 97.6      & 95.4     & 77.2 \\
SSRAN & \ \textbf{77.9}$\uparrow$ & \ \textbf{89.4}$\uparrow$ & \ \textbf{48.9}$\uparrow$ & \ \textbf{98.4}$\uparrow$ & \ \textbf{95.8}$\uparrow$ & \ \textbf{77.5}$\uparrow$ \\
\hline
\end{tabular}
\caption{SLU performance on MixATIS and MixSNIPS datasets.
The numbers with $\uparrow$ indicate that the improvement of our model over all baselines is statistically significant with $p < 0.05$ under t-test.}
\label{res0}
\end{table*}

\paragraph{Baselines:}
We compare our model with the existing baselines, including:
\begin{itemize}
\item Stack-Propagation \cite{qin2019stack}: A LSTM based model with stack-propagation framework and token-level ID.
\item Joint Multiple ID-SF \cite{gangadharaiah2019joint}: A multi-task framework with the slot-gated mechanism for multiple intent detection and slot filling.
\item AG-IF \cite{qin2020agif}: A LSTM-based joint model with an adaptive interaction network.
\item GL-GIN \cite{qin2021glgin}: A LSTM-based joint model with a Global-Locally Graph Interaction Network, which is the current SOTA Multi-SLU model. 
\item LR-Transformer \cite{cheng2021effective}: A Transformer based non-autoregressive model with a layer-refined mechanism. 
\item Basic Model \cite{Attention_is_all_you_need}: A six-layers Transformer Encoder, which is equivalent to only utilizing the encoder and decoder of our model. 
\end{itemize}

\subsection{Result and Analysis}
In this section, we show the results of our experiments and do some analysis.

The experiment results of our model are shown in Table \ref{res0}. Our model significantly outperforms all the baselines and achieves the best performance in all three metrics. Compared with the SOTA baseline \texttt{GL-GIN}, our model enhances the performance by 1.6\%(ID), 1.1\%(SF), and 5.4\%(Overall) on MixATIS, and 2.8\%(ID), 0.9\%(SF), and 2.1\%(Overall) on MixSNIPS. Compared with the prior Transformer-based model \texttt{LR-Transformer}, we enhance the performance by 1.8\%(ID), 1.4\%(SF), and 5.6\%(Overall) on MixATIS, and 2.8\%(ID), 1.4\%(SF), and 2.6\%(Overall) on MixSNIPS.
These results verify the effectiveness of our model intuitively.

Notably, our basic model performs worse than both \texttt{GL-GIN} and \texttt{LR-Transformer}, but \texttt{SSRAN} outperforms both of them, especially on overall accuracy (5.4\% on MixATIS). We attribute this enhancement to the fact that SR captures the scope information, bringing more accurate information to each token. And RAN incorporates the result information of both SF and ID into the model, which mitigates the error propagation effectively. 


\begin{figure*}[t]
\centering
\includegraphics[width=0.95\textwidth]{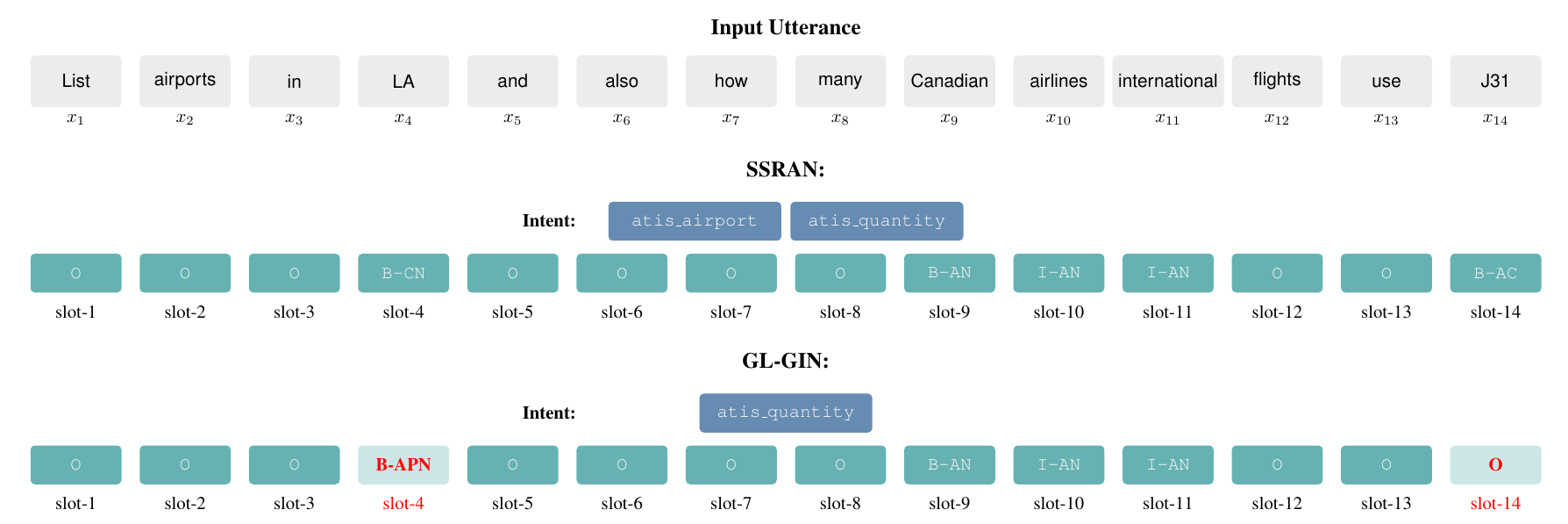}
\caption{Case study for Multi-Intent SLU, where \texttt{CN} indicates \texttt{city\_name}, \texttt{AN} indicates \texttt{airline\_name}, \texttt{AC} indicates \texttt{aircraft\_code}
and \texttt{APN} indicates \texttt{airport\_name}. The slot with red color indicates error-predicted slot label.}
\label{fig: case study}
\end{figure*}

\subsection{Ablation Study}
\label{sec: Ablation Study}
In this section, we do an ablation study to verify the effectiveness of SR, RAN and our auxiliary task in detail. The result is shown in Table \ref{res0}.

\paragraph{Effect of Scope Recognizer:}
We first remove our SR component to verify its effectiveness, which is referred to as \texttt{SSRAN w/o SR} in Table \ref{res0}. Without SR, the performance drops by 1.4\%(ID), 0.7\%(SF), and 2.6\%(Overall) on MixATIS and 1.5\%(ID), 0.4\%(SF), and 0.8\%(Overall) on MixSNIPS. This phenomenon indicates that scope information influences the accuracy of Multi-Intent SLU, especially on the ID task. 

Moreover, as shown in Table \ref{res0}, although both our model and \texttt{LR-Transformer} (directly adding result-embedding to hidden states) consider the bidirectional interaction, \texttt{SSRAN} performs much better than \texttt{LR-Transformer}. We attribute this to the ability of SR to reduce the distraction of the result information outside the scope.

\paragraph{Effect of Result Attention Network:}
To verify the effect of RAN, we remove our RAN component, which is referred to as texttt{SSRAN w/o RAN} in Table \ref{res0}. Without RAN, the slot f1 score drops by 1.8\% (MixATIS) and 1.4\% (MixSNIPS),  and the overall accuracy drops by 5.9\% and 2.4\%, demonstrating the significance of result information.

Note that our model outperforms \texttt{GL-GIN} (only considering ID results in SF) in all metrics, even only utilizing RAN (i.e., \texttt{SSRAN w/o SR}). We attribute this to the fact that RAN considers the bidirectional interaction (also uses SF result in ID), which mitigates the error propagation problem caused by error ID labels.

\paragraph{Effect of Auxiliary Task:}
We finally conduct experiments to find out the effect of the auxiliary task. The results in Table \ref{res0} show that the performance of all three metrics drops on both datasets,
indicating that our auxiliary tasks further improve our model by multi-task learning.
We attribute this improvement to the fact that SCT obtains the sequence dependency and handles the uncoordinated slot problem to help the SF prediction, and the INP task captures ID number information in the utterance level to help the ID prediction.
\begin{table}[h]
\centering
\begin{tabular}{l|c|c}
\hline
\textbf{Model} &\textbf{MixATIS} &\textbf{MixSNIPS} \\
\hline
GL-GIN   & 76.3  & 95.6 \\
\hline
SSRAN (Threshold) & 77.1      & 97.6 \\
SSRAN ($\mathrm{TopK}$-INP) & 77.9  & 98.4\\
SSRAN ($\mathrm{TopK}$-True) & 78.1 & 98.5\\
\hline
\end{tabular}
\caption{ID performance for different predicting methods on MixATIS and MixSNIPS datasets.}
\label{resID}
\end{table}

\paragraph{Effect of Topk Method:}
Moreover, the SOTA baseline \texttt{GL-GIN} utilizes a manually set threshold to select the predicted intent labels, which is different from ours. Thus, we conduct the experiments to show the performance of our model when utilizing threshold, $\mathrm{Topk}$ with INP result, and $\mathrm{Topk}$ with the ground-truth intent numbers, to make a more fair comparison. The results are shown in Table \ref{resID}, where we have the following observations: First, our model still performs better than \texttt{GL-GIN} on ID (0.8\% on MixATIS and 2.0\% on MixSNIPS) even utilizing threshold, which indicates the effectiveness of our model. Second, compared with selecting ID results by the threshold, utilizing the $\mathrm{Topk}$ method further enhances the ID accuracy, which is even close to the ground truth.

\subsection{Case study and Visualization}
We provide a case study from the MixATIS dataset to compare the SLU results generated by GL-GIN and our model, where the input utterance consists of two sub-utterances separated by the token \texttt{and}. As shown in Figure \ref{fig: case study}, GL-GIN misses the intent \texttt{atis\_airport} due to lacking scope information and further incorrectly predicts the slot of the token \texttt{LA} due to the error propagation. In contrast, SSRAN predicts all the slot labels correctly. We attribute this to the fact that our model captures the scope information by SR and effectively mitigates error propagation by RAN, resulting in a more accurate prediction.

\begin{figure}[h]
\centering
\includegraphics[width=0.45\textwidth]{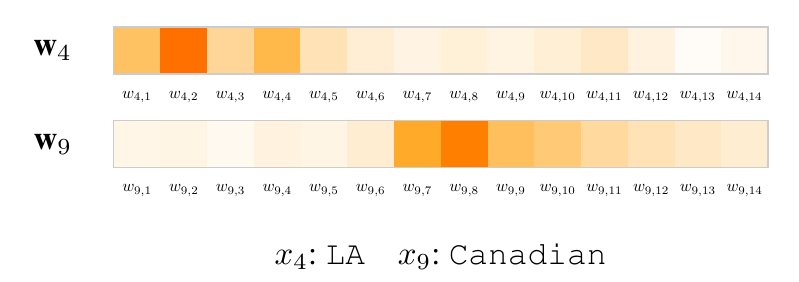}
\caption{Visualization for scope weight matrix $\textbf{W}$ in SR. For each column, the darker the color,
the bigger the weight.}
\label{fig: SR-example}
\end{figure}

Moreover, we provide visualization for the scope weight matrix computed by SR. As shown in Figure \ref{fig: SR-example}, token \texttt{LA} will pay more attention to the information from the first sub-utterance, especially from the token \texttt{airport}, and predict the intent \texttt{atis\_airport}.
On the contrary, \texttt{Canadian} will pay more attention to the second sub-utterance and predict the intent \texttt{atis\_quantity} with the information from the tokens \texttt{how} and \texttt{many}.

\subsection{Effect of Pretraining}
In this section, we conduct experiments to evaluate the ability of our model to combine pretrained models by replacing our encoder with RoBERTa \cite{roberta}, BERT \cite{BERT}, and XLNet \cite{xlnet}, respectively. 

\begin{table}[h]
\centering
\begin{tabular}{l|c|c}
\hline
\textbf{Model}  & \textbf{MixATIS}  & \textbf{MixSNIPS} \\ 
\hline
RoBERTa   & 49.7 & 80.2 \\
AG-IF + RoBERTa   & 50.0  & 80.7 \\
GL-GIN + RoBERTa   & 53.6  & 82.6 \\
SSRAN + RoBERTa  & 54.4  & 83.1 \\
\hline
BERT  & 51.6  & 83.0 \\
SSRAN + BERT  & 54.8  & 84.5 \\
\hline
XLNet  & 52.1  & 84.8 \\
SSRAN + XLNet  & 55.3  & 85.6 \\
\hline
\end{tabular}
\caption{Overall performance for pre-trained models on MixATIS and MixSNIPS datasets.}
\label{resPR}
\end{table}

The experiment results are shown in Table \ref{resPR}, where we have two observations. First, our model is benefited from all three pre-train models, which verifies the compatibility between our model and the pre-trained model. Second, \texttt{SSRAN} outperforms \texttt{AG-IF} and \texttt{GL-GIN} when utilizing RoBERTa, which further verifies the effectiveness of our model on Mult-Intent SLU tasks.

\section{Conclusion}
In this paper, we proposed a scope-sensitive model SSRAN for Multi-Intent Spoken Language Understating task, consisting of a Scope Recognizer and a Result Attention Network. SR focuses on the scope of the intent and guides the prediction by emphasizing closely related information and reducing the distraction of out-of-scope tokens.
RAN effectively mitigates the error propagation problem by utilizing the bidirectional interaction between ID and SF.
Experimental results show our model achieves the best performance on two public datasets.

\section*{Acknowledgements}
This work was supported in part by National Key R\&D Program of China (2022YFE0201400); 
in part by Guangdong Key Lab of AI and Multi-modal Data Processing, United International College (UIC), Zhuhai under Grant No. 2020KSYS007 sponsored by Guangdong Provincial Department of Education;
in part by the Chinese National Research Fund (NSFC) under Grant Nos. 62272050, 61872239; in part by Institute of Artificial Intelligence and Future Networks (BNU-Zhuhai) and Engineering Center of AI and Future Education, Guangdong Provincial Department of Science and Technology, China;
Zhuhai Science-Tech Innovation Bureau under Grant Nos: ZH22017001210119PWC and 28712217900001, and in part by the Interdisciplinary Intelligence SuperComputer Center of Beijing Normal University (Zhuhai).

\bibliography{aaai23.bib}




\end{document}